# The Mutations of Machine Speech

Mauricio Figueroa[1]

(1) Assistant Professor, Durham University
Mauricio Figueroa
Email: mauricio.figueroa@durham.ac.uk

## Abstract

Algorithmic outputs now populate the digital environments through which contemporary life is organized. The role of law in facilitating and constituting (rather than merely responding to) these processes is gaining increasing traction across scholarly accounts. This inquiry traces the evolution of algorithmic outputs attending to their legal underpinnings and social implications, surfacing the "mutations" of machine speech.

The first mutation redefined speech as data to be queried: search engines transformed the web from a space of information retrieval into an economic regime of algorithmic visibility. The second mutation reframed speech as engagement: social media platforms fused moderation with amplification, turning expression into a metric of attention, governed by corporate architectures. The third mutation emerges in conversational systems and interfaces, where generative text displaces information retrieval, bringing with it dense technolegal entanglements and profound epistemic consequences.

Scholars of freedom of expression, informational privacy, and communication studies have long grappled with these dynamics, yet their implications for broader legal thought have also become urgent. This piece seeks to organize and clarify the evolving debate around algorithmic speech, making this critical but often fragmented discourse more accessible to wider legal and interdisciplinary audiences. In doing so, it bridges the gap between observing technological transformation and critically assessing the constitutive role of law within it, offering a conceptual resource for researchers, students, policymakers, and practitioners navigating and contesting this evolving landscape.

## Keywords

# Introduction

Law does not simply react to technological developments; it helps to constitute them. As Cohen (2019) argues in her account of the information economy, legal arrangements actively shape the conditions under which information and value are produced and circulated. Similarly, Jasanoff's (1995) influential account of the co-production of law, science, and technology illuminates the role of law in the social construction and meaning-making of technological systems. This piece maps the evolution of algorithmic outputs beyond their technical operations, attending to what it calls their *mutations*: different transformations in the ways algorithmic speech is produced, mediated, or governed. Each mutation marks a moment when new sociotechnical practices take hold, with different legal configurations sustaining them. This analysis unfolds in three movements, each attending to a distinct mutation, before turning to a synthesis that gathers their insights and a set of concluding reflections. The central claim advanced in this piece is that these mutations can be understood as constituting a broader cartography of the constant role of law in the construction of machine speech.

The first mutation redefined speech as data to be queried. Search engines transformed the web from indexation into an economic regime of algorithmic visibility. Ranking and optimization became the new languages of information. Here, the law plays two different roles: in the United States, free speech doctrine framed search results as expressive acts; in Europe, data protection law framed search results as data practices, subject to the exercise of rights against them. The tension between these approaches reveals law's constitutive function and how it configures the epistemic boundaries of what counts as *speech* and whose speech it is.

The second mutation reframed speech as engagement. Social media platforms fused moderation and amplification, transforming expression into a metric of attention and emotional resonance. Unlike the relatively passive search engine, social media thrives on the active and interactive participation of users, whose content fuels visibility and profit. Within this architecture, law operates both inside and outside: internalized through corporate content policies and risk management, and externally through legislative interventions that attempt to reshape these systems from the outside.

The third mutation emerges with the rise of generative artificial intelligence, where speech becomes dialogical and synthetic. Here, language models no longer retrieve information but produce it, generating text that simulates human communication while remapping the boundaries between authenticity, authority, and persuasion. Upstream, questions of data sourcing, ownership or consent collide; downstream, issues of misinformation, accountability, and even end-user harms proliferate. Law once again stands within the frame, allowing machine speech to attain its form and force.

The goal of this contribution is to offer a way of seeing familiar terrains anew. While the field has advanced through rich and diverse contributions, their density and scope can obscure their interconnections. This chapter seeks to bring those strands into conversation, framing them through the lens of the mutations of machine speech. By mapping the mutations of digital speech, this chapter invites a broader conversation across disciplines, encouraging legal scholars, sociologists, technologists, and political scientists alike to engage with the ways language, power, or legality are being reconstituted in the age of machine communication.

## From Human to Algorithmic Speech

Communication is hardly a capacity unique to the human species, yet the modalities through which humans enact it are distinctive in degree and in kind. Indeed, accounts in anthropology and archaeology underscore how the evolution of language, tool use, and collective life unfolded in constant interaction. Speech was shaped by tools as much as tools were shaped by speech (see Leroi-Gourhan 1993). Thus, the residual belief in an original, unmediated human subject detached from the technical results in a highly contested (and arguably unpersuasive) construct.

The emergence and adoption of digital technologies have reconfigured the architecture of human communication, enabling new modalities of interaction that are instantaneous, disembodied, or persistently available. Digital technologies collapse geographic barriers, enabling real-time engagement across vast distances (see, generally, Floridi (2014)). They sustain remote education, facilitate decentralized forms of labor, and underpin new modes of commerce, cultural production and consumption, while also expanding the horizons of individual expression and political participation. Yet to frame these changes solely in terms of enhanced reach and efficiency is to largely miss their deeper significance. In other words, thinking of digital technologies as only helping humans to do what they had always done (only faster, wider, and with fewer material constraints) misses the mark. Underneath, there is a rather profound transformation in the nature of communication itself.

The line between mediation and communication, between method and speech, has grown porous. What distinguishes the present moment is not necessarily the fact that digital systems mediate human communication. It is the fact that algorithmic systems intervene in the construction of communication and meaning-making by themselves. Search engines, recommendation algorithms, and generative models participate in the very circuits of meaning-making that we once thought to belong exclusively to human interlocutors. Search engines speak in the grammar of relevance; the grammar of social media is that of amplification; and generative models use the grammar of statistical persuasion. Each, in its own way, performs a kind of speech act. To speak today is to do so through interfaces and digital architectures that also communicate by themselves. As a brief aside, it is interesting and telling to note the common practice of listing programming languages alongside natural languages in technical CVs, a convention that implicitly recognizes their communicative character.

Across disciplines, theorists have grappled with how such algorithmic activity should be understood, from a legal standpoint and beyond. The following section surveys these debates and their implications, tracing how or when algorithmic power acquires the status of speech and examining the role that legal structures play in this process. In particular, we will unpack how search engines, social media platforms, and generative systems each instantiate distinct mutations, shaped by underlying business models and by the mobilization of legal frameworks that enable and constrain them.

## Law Shaping Algorithmic Speech

The preceding section departed from the premise that humans are not the only agents operating within the digital spectrum and that computational systems have become crucial actors in shaping, ordering, or even producing the communicative landscape. Whereas the notion of humans employing online tools to communicate, organize, or express themselves is intuitively grasped through everyday experience, particularly by wider audiences, the disciplining and ordering power of algorithms (and their emergent expressive force) has become a central concern of influential and contemporary scholarship (see, generally, Gillespie (2018), Kaminski and Jones (2023), Miller (2021), Bietti (2023), and Cohen (2023))

Indeed, sustained scholarly engagement has sought to map the terrain of algorithmic expression, exploring how such outputs might be interrogated as communicative acts and where, within that terrain, law's authority and influence are to be situated. As Margot Kaminski and Meg Leta Jones highlight with remarkable clarity, it is not the case that law has stood on the sidelines, passively reacting to new forms of digital speech as they arise. Rather, legal institutions have actively participated in defining what counts as "speech" in these contexts, determining the scope of the protections or obligations such classifications entail (Kaminski and Jones 2023). In doing so, law has shaped not only the politics of algorithmic systems but also the very meaning of their outputs. The following is one attempt to organize and classify the different mutations of digital speech and the roles of law in shaping each of them.

## From Indexation to Ordering

The Internet, which began as a governmental project and soon evolved into a domain of private experimentation, introduced novel possibilities and imaginaries of informational life. The aspiration to make information universally accessible (anywhere, anytime, through a personal computer) set off a series of experiments in information retrieval. These crystallized with the World Wide Web, which promised both new modes of access and new architectures of visibility (Abbate 2000; Leiner et al. 2009).

By the late 1990s, search engines presented two distinct modalities, both of which look strikingly alien when measured against today's default pathways for navigating the web. Some curated directories of links (early Yahoo), while others matched strings of keywords (AltaVista) (Schwartz 1998). These arrangements would not last long. The turn of the millennium marked a decisive shift. Google rose to prominence by introducing *PageRank*, an iterative algorithm that evaluated web pages not merely by their textual content but by their position within a network of hyperlinks. (See Bianchini et al. (2005) for an explanation of page authority vs page content.) This means that once a page or website is linked or referred to in other sites, its relevance increases. On the assumption that valuable pages are those most frequently and authoritatively linked, Google recast retrieval as an exercise of calculating relational significance.

The move to sort the web in line with hyperlinking metrics was innovative but also met with early and boundary-pushing scholarly critique. As Helen Nissenbaum and Lucas Introna observed, because ranking systems are never neutral, commercial content would tend to displace non-commercial domains. (See Introna and Nissenbaum (2000), where the authors argue with remarkable precision how systematic inclusions and exclusions in search results are indeed political issues, because they determine what people (the seekers) are able to see and, perhaps more importantly, what the web represents for them. I have elsewhere provided additional account regarding the visible area vs scrolling area and its legal implications. See Figueroa-Torres (2023).)

Furthermore, at least three further elements contributed to the paradigm shift of web search. First, the introduction of advertising schemes reshaped the search engine into an economic platform: Google incorporated ads directly into search results, enabling businesses to bid for keywords and purchase visibility (Lee 2011). Second, personalization, introduced in 2004 and made universal by 2009, brought user-specific variables (such as language, location, or search history) into the ordering process (Hannak et al. 2013). Third, the rise of the Search Engine Optimization (SEO) industry illustrated the growing centrality of search engines to commerce and spurred strategic manipulation of ranking criteria (Wu and Davison 2005). Practices such as link farming (i.e., densely interlinked networks of websites designed to inflate relevance) compelled Google to demarcate "organic" from "spam" results, establishing hierarchies of visibility and legitimacy on the web, giving rise to the familiar difference between sponsored and organic results in web searches.

Search results became more than the mechanics of retrieval and mathematical calculation. They reconfigured visibility as a scarce resource to be bought, not just found. Personalization, in turn, made that the same query would result in diverging results for different individuals, opening up the debate on what results are being shown, but also which possibilities of encounter are foreclosed. Anti-spam interventions, in turn, endowed platforms with a perceived legitimacy to demarcate synthetic from authentic relevance. (See Boyd and Golebiewski (2018) for a deeper explanation of how malicious actors manipulate search engine results.)

This transformation has not gone uncontested. As search engines assumed responsibility for determining what content would be rendered visible or consigned to obscurity, their growing authority became the subject of public, scholarly, or legal debate. Courts, in particular, were pressed to decide whether the ranking of search results should be understood as falling within the ambit of freedom of expression. In *Search King, Inc. v. Google Technology Inc.* (2003), the US Supreme Court concluded that Google's *PageRank* constituted an opinion about the relative significance of websites rather than a factual assertion and was therefore protected under free speech law. In so holding, the Court put *PageRank* in the vicinity of editorial judgment, similar to that exercised by newspapers in selecting and arranging content.

The recognition of *PageRank* outputs as constitutionally protected speech did more than resolve a dispute between two firms. It co-created the status of algorithmic assessments of relevance as legitimate acts of editorial discretion. In turn, this has added a further layer to what is a *search engine* ontologically and legally speaking. It is necessary to point out that a search engine is a complex legal creature. At once a proprietary computational system, safeguarded by intellectual property (IP) law, and a speaker whose outputs enjoy free speech protection. The coexistence of IP and free speech protections is not novel; films and novels, for example, have long been shielded by both regimes, even when challenged under defamation law. (Copyright has safeguarded their authors' economic and moral rights, while freedom of expression has defended their communicative and artistic value when challenged under defamation or censorship law. This is attached, of course, to the publicness or notoriety of the person depicted.) What is distinctive in the case of search engines is the way their infrastructural capacity intersects with the evolving governance of digital speech. Search engines are cast simultaneously as bearers of free speech protection and as infrastructural actors that shape the conditions of expression for others.

This hybrid positioning speaks to (but does not map neatly onto) Jack Balkin's seminal account of "old school" and "new school" speech regulation (Balkin 2013). In Balkin's triangular schema, one corner is occupied by state authorities; another by speakers, such as end users and legacy media; and a third by digital intermediaries such as social media platforms (Balkin 2018). Old school regulation describes the direct interventions of state actors in relation to speakers such as journalists. New school regulation captures the entanglement of state authorities with digital infrastructure providers, in which collaboration or co-optation enables the governance of speech through private platforms. A third set of relations, between intermediaries and users, consists of forms of private ordering through contractual terms, content policies, or algorithmic curation. Search engines complicate this geometry. They do operate as infrastructures through which speech flows; but their outputs (their rationale of existence) enjoy the privileges of protected speech as well. This hybrid position (simultaneously infrastructural mediators of speech with protected speech outputs in their own) pushes against the limits of the triangle, exposing the ways in which infrastructural power can be naturalized as expressive freedom. What matters here is not merely that a corporate entity enjoys speech rights. A corporation may easily endorse a political candidate, donate to political causes, or take public stances on cultural and social issues. Here, it is the technical operations of an algorithmic system (indexing, ranking, or curating) that are constructed as constitutionally protected expression (Volokh and Falk 2011).

Because machine outputs have no intentional *speaker* in the human sense, contesting them becomes less about rebutting a speaker than about challenging their status as "speech". As Karminski and Jones point out, the doctrinal threshold question is not whether there is a *speaker* but whether there is *speech* (Kaminski and Jones 2023). That, precisely, is what the jurisprudence of algorithmic speech accomplishes: it endows computational processes with the legal affordances of expression. Perhaps a way to understand a search engine within Balkin's triangular approach, is to see it as a moving part, meaning its position in the triangle *shifts* depending on whether it acts as a system whose speech is to be protected or intermediaries to be regulated.

The framing under American law of search engines outputs is well rooted under the paradigm of freedom of expression. That framing, though, is not necessarily universal. In the EU, freedom of expression is important but balanced against other rights (e.g., privacy or non-discrimination). The grammar through which search engines are interrogated in terms of their outputs in the EU is, to a good extent, that of informational privacy or data protection. For instance, in the now well-known case of *Google Spain v AEPD and Mario Costeja González*, the Court of Justice of the European Union held that search engine operators are responsible for the processing of personal data appearing on web pages published by third parties (*Google Spain SL and Google Inc. v. AEPD and Costeja González*, 2014). This, in turn, derives in the fact that an individual may exercise the right to deletion against the indexation of the search engine, entailing the erasure of that link and altering the results. The legal configuration that comes up is not necessarily the naturalization of algorithmic processes as speech but their juridical construction as data practices, subject to data rights, including rectification and deletion. The contrasting choices to treat indexing as speech or as processing is precisely a defining element of machine speech's mutating nature and how legal grammars actively configure the boundaries of technology.

Similarly, the Digital Services Act (DSA) frames very large online search engines (VLOSEs) as entities subject to specific duties. They must conduct risk assessments, allow audits, provide access to vetted researchers, among other duties. Their legal obligations flow from their role as infrastructures of visibility and their character of speechmakers seems to be largely outside of the equation when compared against their duties to control and monitor the amplification of content (Husovec 2023). Importantly, the DSA even explicitly links to General Data Protection Regulation (GDPR) obligations, reinforcing the idea that indexing and ranking are regulatory targets, not constitutionalized expressions (DSA, 2022, arts. 25, 38).

The framing of search results as speech, on the one hand, or as data practices, on the other, has consequences that reach well beyond abstract doctrinal debates. In jurisdictions of the Global South, courts and policymakers struggle not only with how to categorize search engines but also with how to situate them in their legal and institutional traditions. In the pending case *Ulrich Richter v. Google Mexico* (*Richter v Google Mexico*, 587/2017 (pending)), for example, the Mexican Supreme Court is now confronted with the question of how to accommodate a defamation claim within a civil-code tradition while facing pressure from civil society groups. They filed amicus curiae framed in the familiar language of non-intermediary liability and freedom of expression (Wikimedia Foundation 2024), but leave largely unexamined the asymmetrical power of search engines over their users and the striking absence of remedies or procedural avenues through which individuals might contest reputational harm. The anxieties to preserve the status quo mirror the liberalism and techno-solutionism discourses of the 1990s that Bietti has identify in her mapping of platform regulation (Bietti 2023). To their credit, civil society organizations aligned with (and at times funded by) Big Tech have defended pluralism online. But they continue to inhabit an outdated vision of cyberspace as an independent domain beyond political intervention. The catastrophes they predict are overstated; the promises attached to legal victories against Big Tech are, likewise, exaggerated.

Elsewhere in the region, Argentina has developed a more substantial body of case law against indexation practices. In *Carrozo v. Yahoo de Argentina* (Van Calster et al. 2018), the judiciary held both Google and Yahoo liable for the dissemination of non-consensual sexual images. The damages awarded were relatively modest and it remains unclear whether those decisions produced meaningful changes in search engine mechanics, comparable to the GDPR-related adjustments implemented in Europe in response to delisting requests. Even so, Argentina occupies a distinctive position within the Global South as a jurisdiction in which judicial challenges to search engine power have been pursued with unusual persistence (Carter 2013).

## From Moderation to Amplification

While the previous section dealt with how search engines index and mediate access to existing information, this section now turns to analyze how users moved from being consumers of information to producers of content, capable of contributing to and reshaping the informational landscape, and the algorithmic consequences thereof. Scholarly accounts have advanced how Social Networking Sites (SNSs) allowed users to create and disseminate content; but more importantly, they restructured the conditions under which such creation could occur (Gillespie 2018). Platforms supplied the technical affordances that rendered content generation and circulation accessible to the public within commercial architectures of mediation and control.

In the early stages of the commercial Internet, the proliferation of user-generated content emerged alongside the imperative for its moderation. People uploaded content where it could be seen, whether their messages caused laugh, inspired reflection or triggered offense. Social media intensified contact among strangers and widened the scope of interaction, organizing users into new, networked publics (Gillespie 2018). In an influential and well-documented work, including personal interviews with relevant actors in the industry, Kate Klonick highlights how in their origins many online commercial platforms saw themselves as online services or software companies rather than *speech* platforms (Klonick 2017). Chats, bulletin boards, or user posts appeared more as peripheral features within a larger constellation of hosting and subscription offerings than as core sites of public discourse requiring oversight. The idea that these companies might actively police user-generated expression (let alone do so in a consistent or coordinated manner) was largely absent from their initial self-conception. As user-generated content grew, in-house lawyers, in their origins quite small, help to formalize content moderation policies inspired by an underlying influence of free speech law, filtered through corporate self-interest (Klonick 2017). Some issues lent themselves to ready-made procedural models. Copyright claims, for instance, could be channeled into the notice-and-takedown system codified in the U.S. Digital Millennium Copyright Act (DMCA). A rights-holder notified the platform of infringement; the platform took the content down, while retaining its *safe harbor* from liability (DMCA, 1998, § 512).

Whereas copyright serves as a register to govern content of its province, not every dispute concerned intellectual ownership. Offensive, hateful, or harmful content required judgment; i.e., an assessment of meaning, context, and nature. Unlike copyright enforcement, there was no proprietary rights-holder as such, no straightforward rule of action. This pushed platforms to develop rules of their own. Public-facing community guidelines offered a simplified account of permissible conduct. Internally, more detailed manuals instructed moderators (human and algorithmic alike) how to decide what stayed up and what came down. Over time, ad hoc practices hardened into policies, policies into infrastructure, leading into automation and human-review configurations.

American legal scholarship has consistently and extensively mapped this terrain, interrogating the promises and drawbacks of ex-ante rules and ex-post adjudication (see, generally, Klonick (2022)), the alternatives for moderation design and scale, and the corporate interests embedded throughout. The field is rich yet never settled.

Alongside moderation, another question that has gained urgency is how content travels. The issue is no longer just what platforms remove, but what they amplify. Algorithms do not, or at least not only, host expression, they provide a platform for content to reach different audiences and to scale up. Familiar sayings like content "goes viral" evoke contagion but might dilute the commercial, attention-seeking imperatives that drive that viralization. Whereas the previous mutation confronts us to the question of speech in the absence of speakers, this mutation makes us appreciate that it is not what the users say that really matters, but how platforms choose to amplify and monetize.

As Tarleton Gillispie highlights, early coverage focused precisely on moderation and isolated missteps, where decisions appeared rather incohesive (Gillespie 2018). Today, contemporary debates compel us to analyze the overreach and censorship around content, the derived harms left unchecked, or the hidden labor force (largely outsourced) that sustains content moderation systems today.

Legally, it is tempting to treat moderation and amplification as distinct. Yet in practice, they intertwine. Both are products of the same sociotechnical systems and the same legal architectures. Two US statutory frameworks sustain this logic: the Communications Decency Act (CDA) and the Digital Millennium Copyright Act (CDA). The CDA's Section 230 shields intermediaries from liability for user speech, casting them as neutral carriers. The DMCA, by contrast, imposes duties of removal when intellectual property rights are invoked. Together, they built one major part of the legal scaffolding of the platform economy. Both have been deeply internalized in corporate governance and design. Yet both remain structurally indifferent to the realities of scaled, algorithmically mediated speech.

Erin Miller describes the relationship between content amplification and free speech doctrine in the US constitutional law as one of *amplification blindness* (Miller 2021). The prevailing paradigm holds that speech is protected regardless of its reach. Protection thus shields speakers only from state interference, leaving access to amplifying mechanisms to be determined by private ordering. At the same time, Miller argues, case law has consistently favored editorial autonomy of those who control such mechanisms, grounded in case law protecting printed media newspapers against statutory obligations to print opposing views. The cumulative effect, Miller suggests, is the naturalization of disparities in communicative power: those who command the most potent amplifiers retain broad, largely unchecked authority over which voices enter public discourse and on what terms (Miller 2021).

Here, then, lies the second mutation. Amplification becomes algorithmic; the algorithm, in turn, stands beside law, quietly co-governing what is seen and scaled. Its operations are not directly commanded by law but authorized through its silence; made legitimate by what law chooses not to forbid.

Researchers in the field of media and communication remind us that algorithmic amplification mechanisms do not merely respond to user preferences; they actively structure them (Huszár et al. 2022; Mansoury et al. 2020). As these recommendation systems have grown more sophisticated, the flow of content has become less a product of intentional choice and more the outcome of opaque, continuously adaptive processes designed to maximize engagement.

Critics warn that these dynamics amplify bias, privilege engagement over epistemic diversity, exceeding the understanding of their own designers and raising profound concerns about autonomy, democratic deliberation, or the circulation of extremist content (Whittaker et al. 2021). The need for ongoing and wider empirical research on this issue is a pressing matter, yet the access to information via APIs has met substantial complications in recent years, with one major technology platform closing its access to academics for independent scrutiny in the aftermath of the COVID pandemic (Murtfeldt et al. 2024).

The stakes of amplification, however, are not limited to the United States or Europe. They become existential in places where linguistic marginalization and fragile civic institutions meet. The Global South demands a sharper lens.

In Myanmar, the absence of effective moderation, exacerbated by the scarcity of Burmese-language moderators and the platform's relegation of the country to a "second-tier" market, allowed inflammatory speech to circulate with minimal friction against a Muslim segment of the population (Amnesty International 2022). Engagement-driven algorithms did not merely mirror existing tensions; they operationalized them. The logic of attention extraction rewarded inflammatory and emotive content, privileging division and resentment.

To be sure, the atrocities that followed were committed by people, not code. Yet it would be a mistake to imagine that algorithmic systems stood apart. By amplifying hate-filled content and normalizing its presence in everyday feeds, these systems helped shape a public atmosphere in which violence could be imagined (and materialized) as collective action.

If companies insist that a funny cat video and a video inciting hatred toward a religious minority can travel equally easily through their systems (so long as both attract attention), they reveal a fracture in their own story, their own professed norms and alleged corporate values.

Once, moderation and amplification seemed worlds apart: one about taking speech down, the other about sending it farther. Today, that line has become blurred. Both shape the public sphere; both exercise power through design. Now, policy debates analyze them in tandem. Significantly, European regulators have begun to address them together, calling for transparency and oversight. Whether these interventions will suffice, or must be reimagined, remains an open and deeply contested question (Cohen 2023). This is particularly manifested in the EU's DSA.

The DSA establishes a comprehensive regulatory framework for online intermediaries and platforms, including social networks, content-sharing services, digital travel and accommodation providers, among others. At its core, the DSA is animated by a dual imperative: to curtail the proliferation of illegal and harmful content, including disinformation, and to reshape the digital environment in ways that safeguard users' safety, and fundamental rights. In that form, the DSA gestures toward a deeper transformation: the governance of amplification itself. In this sense, the DSA is an attempt to make visible and govern the infrastructures of algorithmic speech, the relatively quiet structures through which attention is captured, redistributed, monetized or scaled up. Yet, it is early still. Its first years are marked less by resolution than by negotiation, as institutions, auditors, and compliance officers learn how to inhabit (or manipulate) its legal grammar.

The road ahead is not easy. By formalizing accountability through risk assessments, audits, or compliance protocols, the DSA risks reproducing the very bureaucratic opacity it seeks to dismantle. Lessons in the field of data protection remind us that forms, certifications, or procedural rituals may obscure the living dynamics of power and provide a sense of compliance and legitimacy (see, generally, Cohen and Waldman (2023) and Waldman (2021)). Still, the DSA stands as a major consolidated effort to govern amplification, however imperfectly.

As Martin Husovec observes, the DSA represents the first serious attempt to articulate a "second generation" of digital service regulation, one that addresses the complexities of user-generated content ecosystems with a deliberately horizontal, cross-sectoral approach. Its architecture presupposes an active role for both state and non-state actors in sustaining the necessary checks and balances. Some aspects of the regime, he notes, may appear "too European" in their institutional design and normative ambitions. Yet the principles it embodies, especially its insistence on shared responsibility and structural accountability, could inform regulatory experiments beyond Europe's borders, including in the United States (Husovec 2023), where the search for a comparable framework is still unresolved.

To legislate in such an environment is to intervene in the very conditions of communicability: it is to decide what kinds of visibility may flourish and what forms of silence must be sustained. Whereas content moderation implied a set of normative and sociotechnical configurations to determine what is being left out there or taken down, scaled-up speech reveals that harms are not only cumulative in number but transformative in kind, capable of producing profound social and political consequences.

The law has largely acted as participant and translator in the mechanics of algorithmic speech, first through the regimes of moderation, now through the more elusive terrains of amplification. The former, through the grammar of copyright and free speech, structured the normative register of digital governance, particularly in the American context. Amplification, however, it is easily mistaken for an environmental feature of the networked world, "the digital," "the social," or "the platform"—whatever metaphor we summon to make its logics feel inevitable. Whereas individual instances of infringement are rendered as anomalies, amplification is received as the normal weather of communicative systems. As discussed above, this is the kind of blindness that Miller brings into view in her work.

Europe's experiment with the DSA seeks, however imperfectly, to name and govern this speech infrastructure: to render the operations of amplification visible, accountable, or perhaps even disputable. Its success will not really depend on disciplining platforms through procedural ritual but on cultivating new ways of seeing and contesting the architectures of attention that organize public life. Even if the DSA proves to be a missed shot, it is a necessary one, as it is an effort to legislate less on the content of speech as such than in the conditions that bring it about.

## From Information Retrieval to Generation

So far, we have traced two mutations of algorithmic speech. The first, a movement from indexing to ranking (from simple inclusion to hierarchical arrangement) and how that mutation recast the task of search as one of judgment. The second, a mutation not of substitution but of combination: moderation intertwined with amplification, where systems no longer merely remove or suppress but tune and intensify the flows of visibility. Both transformations concern human expression and its algorithmic mediation, as well as the legal architectures that sustain and normalize such arrangements. They are, in other words, technolegal arrangements that configure the conditions of public legibility.

Until now, our analysis has centered on human-generated texts and the algorithmic dynamics that govern their encounter with audiences. What remains less examined is the language produced by machines for human consumption. By *machine-generated speech*, I refer not only to visible rankings or filtered feeds (those, too, are algorithmic utterances) but to the production of text that reads as speech, addressed to us in our own linguistic terms or, in the jargon of computer science, in *natural language*. Here, algorithms do not merely amplify, demote, rank, or retrieve; they mimic human speech as such.

Early conversational systems were retrieval engines at heart. They matched input to template, recombined fragments from a corpus, and returned responses in line with the rules programming them and the pre-set snippets of text (Figueroa-Torres 2024). Generative systems, particularly large language models (LLMs), alter this logic. They do not locate or recombine; they *produce*. By modeling the statistical contours of human language across vast corpora, they generate new sequences of text that, while conditioned by prior expression, may never have existed before in precisely that form (Center for Research on Foundation Models 2021).

Machines that once echoed human language now appear, at least provisionally, to participate in it, though as Emily Bender et al. remind us, their probabilistic foundations and the absence of intentionality that distinguishes them from human speech make them *stochastic parrots* (Bender et al. 2021). Similarly, Wachter et al. introduce the notion of *careless speech*. This is speech unburdened from due regard for truth. They argue that large language models are structurally predisposed to generate such speech. Matters of fact are not determined by appeal to empirical verification but by statistical prevalence within the training corpus. LLMs perform relatively well when questions have clear, widely attested answers that recur frequently in the data. Yet where answers are ambiguous, contested, or time-sensitive, the same probabilistic mechanisms yield responses that merely sound plausible. Its immediate effect is to misinform individual users; its longer-term consequence is the gradual erosion of epistemic confidence and of shared standards for making and testing truth claims (Wachter et al. 2024).

It is against this backdrop that generative systems both depend upon and destabilize the juridical constructs that underwrite them.

There are two specific angles that surface the juridical construct of generative speech. One axis directs attention to the data-extractive architectures through which such speech is produced; the other traces how users meet these machinic utterances as though they were human voices. The analysis that follows can hardly be exhaustive, but it draws upon three interrelated contributions, which illuminate the mutating nature of algorithmic speech and the technolegal configurations that give rise to it. Daniel Solove and Woodrow Hartzog, and separately Alicia Solow-Niederman, direct our attention to the legal terrain upon which generative models emerge (Solove and Hartzog 2025; Solow-Niederman 2026). Margot Kaminski and Meg Leta Jones, in turn, place their attention to the outputs and the role of the law in configuring AI speech (Kaminski and Jones 2023). Together, these works offered a solid terrain to interrogate the law's role in this third mutation of algorithmic speech.

Solove and Hartzog begin at the practice of scraping itself. They argue that mass scraping should be contested and framed as a form of surveillance: an intrusion that undermines informational privacy and security. Solow-Niederman advances how the two primary regimes that govern information flows (privacy law and intellectual property) are collapsing. On the one hand, AI companies justify scraping as the use of "publicly available" information, invoking openness to sidestep privacy concerns. On the other, they fence off the resulting generative models behind proprietary walls, invoking copyright to control access and value. In this inversion, public visibility becomes the pretext for private appropriation. Through their analyses, these scholars render visible how generative systems emerge through a tactical mobilization of legal categories, where the boundaries of privacy and property are eroded or retooled.

Kaminski and Jones shift the frame from inputs to outputs. They distill two legal constructions of AI speech relevant for the understanding of this third mutation. The first, AI as speaker, extends free speech protection to algorithmic expression, effectively rendering machine-generated content as "speech" within constitutional doctrine. This approach raises the paradoxical possibility that outputs produced without intent or accountability may nevertheless receive the law's highest expressive protections, very much as it happens with search engine results. The second construction, drawn from consumer protection law, reframes AI not as a speaker but as a product, or, more provocatively, as a manipulative agent. In this view, generative systems are not autonomous speakers but instruments that distort markets, mislead consumers, or obscure their own mechanisms of influence.

Placed against the previous two mutations, generative systems expose a deeper reconfiguration of the technolegal order. The generative turn signals how the law is stretched and repurposed. Legal interventions in the European Union, such as the Artificial Intelligence Act (AIA), gesture toward accountability of providers of generative models, but leave largely intact the underlying legal regimes through which these systems consolidate power (AIA, Chapter V). The scraped and reordered fragments of the web (related to the prior mutations from indexing to ordering and from moderation to amplification) are carried forward into the generative present.

As human users encounter these systems (and they increasingly do), they face familiar failures of accuracy and misinformation. But additional risks loom. Consider the now-famous case in which a generative chatbot, citing a fabricated news story, falsely accused a law professor of sexual harassment. It placed him at a university he had never taught at, during a trip to Alaska that never happened (Marcus 2024). The harm was real, even if the facts were not. Who is to blame? And can the company behind the model claim the protections of free speech?

The question of free speech and algorithmic outputs landed in the tragic case of a teenager who took his life after interacting with a companion chatbot. When the family sought to hold the company accountable, the provider immediately moved to dismiss the suit, invoking the First Amendment and asserting that the model's outputs constituted protected speech. The federal district court in Orlando declined that invitation, stating pointedly that it was "not prepared to hold that the Character A.I. LLM's output is speech." (See Garcia v Character Technologies, Inc, 20 May 2025 (granting in part and denying in part Motions to Dismiss). Additionally, Gordon-Tapiero (2026) provides a careful examination of the case and She advances a detailed framework of AI liability for artificial companions, relying initially on litigation as targeted legislation evolves. The framing of AI as a product, rather than a service or a speaker, is crucial in her proposed framework. The case, however, is to be settled out of court. (Garcia v Character Technologies, Inc No 6:24-cv-01903-ACC-DCI (MD Fla, 7 January 2026) (Notice of settlement in principle) ('hereby give notice that the Parties have agreed to a mediated settlement in principle to resolve all claims between them in the above-referenced matter').)

This third mutation is predicated on the advancement in technical capabilities of machine speech, but also in the deepening entanglement with the social conditions that give it meaning. Generative systems generate responses so fluent and rhetorically adept that they persuade, seduce, and sometimes manipulate users into taking data-driven patterning for truth. While it may be tempting to see this as a new epoch, it is not necessarily so. It reverberates through earlier moments in the history of computing. It echoes the "computers as social actors" paradigm (Nass et al. 1994; Reeves and Nass 1996) and reaches even further back, with Weizenbaum's discovery that ELIZA could evoke genuine attachment even from users who knew it was a machine (Weizenbaum 1976). What distinguishes this latest mutation is its scale and saturation. The wholesale appropriation of human speech as training material, and its recursive redeployment in machine form. Language that once animated the web as a record of human thought now returns to us as machinic echo, shaping not only the informational environment but also the affective and moral terrains we inhabit (Figueroa-Torres 2024). There will, of course, be space for new modes of co-creation and for forms of knowledge-making that enrich the lived experience of users and perhaps even the social fabric itself. But within these openings, risks also gather, deriving in harms that range from the erosion of privacy and reputational destruction to despair, mental distress, or even suicide.

The frontier of generative speech is also the frontier of law's own imagination. The question is of course the content of what these systems say, but it is equally about what the law allows them to mean.

# Synthesis

Machine speech mutates as technologies evolve, but also because institutions, social practices, and legal regimes continually reorganize the conditions under which communication becomes operational.

Each mutation marks a shift in the architectures that mediate how knowledge is shared and metabolized. The first mutation endows search engines with free speech protection in their outputs. The second mutation recast speech as engagement. Here, social media platforms turned expression into a performative metric, entangling moderation and amplification within an attention economy.

The third, emerging through generative AI, reconfigures speech as dialogue; a continuous, synthetic exchange, producing utterances that both mimic and remake human discourse.

This contribution provided a cartography to organize the shifting architectures through which visibility and legitimacy are distributed. The legal system, too often cast as external to these dynamics, is in fact and internal part of them. Legal doctrines, categories, or silences structure the very field in which algorithmic speech takes form.

Mapping these mutations reveals both continuities and ruptures: datafication, metrics, and commodification persist, but profound reconfigurations of agency and use divert. Algorithmic speech, for instance, in Google Search, Twitter/X, and ChatGPT are not equivalent instantiations; each articulates a distinct relation among data, user, and output and thus a different mode of governing sense-making. To apprehend these differences is to resist the temptation to treat "the algorithmic" as a monolith and instead to view it as a layered, evolving ecology of communicative power.

These mutations I have hereby traced are neither complete nor stable; they are recursive and ongoing. The contribution of this work is therefore diagnostic rather than prescriptive. It opens a terrain of inquiry; one that invites dialogue across disciplines, from law and STS to sociology, linguistics, cognitive science, and philosophy of technology.

# Concluding Remarks

This inquiry, necessarily partial, has traced only some of the pathways through which digital speech mutates. Other routes remain to be mapped. The analysis offered here should therefore be read not as a definitive account but as a provisional cartography, one that marks certain contours and leaves others open for exploration.

It is worth attending to the ways in which the mutations of machine speech are folding in on themselves, with different components that are difficult to disentangle. Search infrastructures now embed large language models within their outputs, while AI-generated content circulates and amplifies within networked social environments. The result is a kind of legal indeterminacy, where outputs resist easy classification, and responsibility becomes harder to locate.

Geopolitics and regulatory variation will inevitably shape how these systems are understood and governed, particularly with respect to the legal status of machine outputs and the conditions under which protection is extended or withheld. Some jurisdictions may impose heightened obligations grounded in public values, while others may be more inclined to accommodate proprietary claims over system-generated works. Meanwhile, large technology firms continue to engage in forms of *legal entrepreneurship*, strategically articulating their own regulatory interests in ways that align with favorable normative regimes (see, generally, Arun (2025) and Cohen (2025)). How these dynamics unfold remains to be seen. There is reason to expect both regulatory fragmentation and ongoing contestation.

The work ahead belongs to a wider community of scholars and practitioners; those willing to ask, once more, what it means to speak, to listen, and to make sense in a world increasingly shaped by machines, or more specifically, machines that attempt to do the same.

Competing Interest Declaration

The author(s) has no competing interests to declare that are relevant to the content of this manuscript.